\documentclass[10pt,twocolumn,letterpaper]{article}

\usepackage{wacv}
\usepackage{times}
\usepackage{epsfig}
\usepackage{graphicx}
\usepackage{amsmath}
\usepackage{amssymb}

\usepackage{multirow}
\usepackage{booktabs}
\usepackage[table,xcdraw]{xcolor}

\usepackage{algorithm}
\usepackage{algorithmic}
\def\wacvPaperID{807} 

\wacvfinalcopy 

\ifwacvfinal
\def\assignedStartPage{1} 
\fi

\ifwacvfinal
\usepackage[breaklinks=true,bookmarks=false]{hyperref}
\else
\usepackage[pagebackref=true,breaklinks=true,colorlinks,bookmarks=false]{hyperref}
\fi

\ifwacvfinal
\else
\fi

\begin{document}


\title{Can the state of relevant neurons in a deep neural networks\\ serve as indicators for detecting adversarial attacks?}

\author{Roger Granda \hspace{1cm} Tinne Tuytelaars\footnotemark[1] \hspace{1cm} Jose Oramas\footnotemark[2]\\
{~}\footnotemark[1]{~}KU Leuven, ESAT-PSI \hspace{1cm} {~}\footnotemark[2]{~}University of Antwerp, imec-IDLab\\
}

\maketitle

\begin{abstract}
We present a method for adversarial attack detection based on the inspection of a sparse set of neurons. We follow the hypothesis that adversarial attacks introduce imperceptible perturbations in the input and that these perturbations change the state of neurons relevant for the concepts modelled by the attacked model. Therefore, monitoring the status of these neurons would enable the detection of adversarial attacks.
Focusing on the image classification task, our method identifies neurons that are relevant for the classes predicted by the model. A deeper qualitative inspection of these sparse set of neurons indicates that their state changes in the presence of adversarial samples. Moreover, quantitative results from our empirical evaluation indicate that our method is capable of recognizing adversarial samples, produced by state-of-the-art attack methods, with comparable accuracy to that of state-of-the-art detectors.
\end{abstract}


\section{Introduction}
\label{sec:intro}

Representations learned via deep neural networks have boosted the performance of artificial intelligence systems on several tasks including image recognition~\cite{HeResNetCVPR15}, object detection~\cite{carion2020endtoend}, speech recognition~\cite{DNNSpeechRecognition19}, text translation~\cite{lewisBARTACL20}, etc. Here, we focus on two main observations over Learning-based representations that have been made in recent years regarding the distribution of the representation that is learned and the sensitivity that this representation may have w.r.t. external attacks.
More specifically, on the one hand, recent research~\cite{frankleLotteryTicketICLR19,hintonDistillingNIPS15,deepCompressionICLR16,oramas2017visual} has shown that despite their extended depth, dense number of internal connections and high amount of parameters, just a small sparse set of neurons are the key elements responsible for the performance obtained by the model.
On the other hand, other research~\cite{szegedy2013intriguing,papernot2016limitations,goodfellow2014explaining,moosavi2016deepfool} have shown that these deep models and learned representations are sensible to small imperceptible perturbations that can be used to make the model behave in an undesirable manner. In the literature, these perturbations are usually referred to as "adversarial attacks".

Taking as starting point the observations from above we formulate the following hypothesis - adversarial attacks are successful at causing unexpected behaviours because the small perturbation they introduce targets the subset of relevant neurons in the model. Therefore, the state of these relevant neurons 
could serve as an indicator to verify whether the model is facing an adversarial attack.

\begin{figure}
  \centering
  \includegraphics[width=0.5\textwidth]{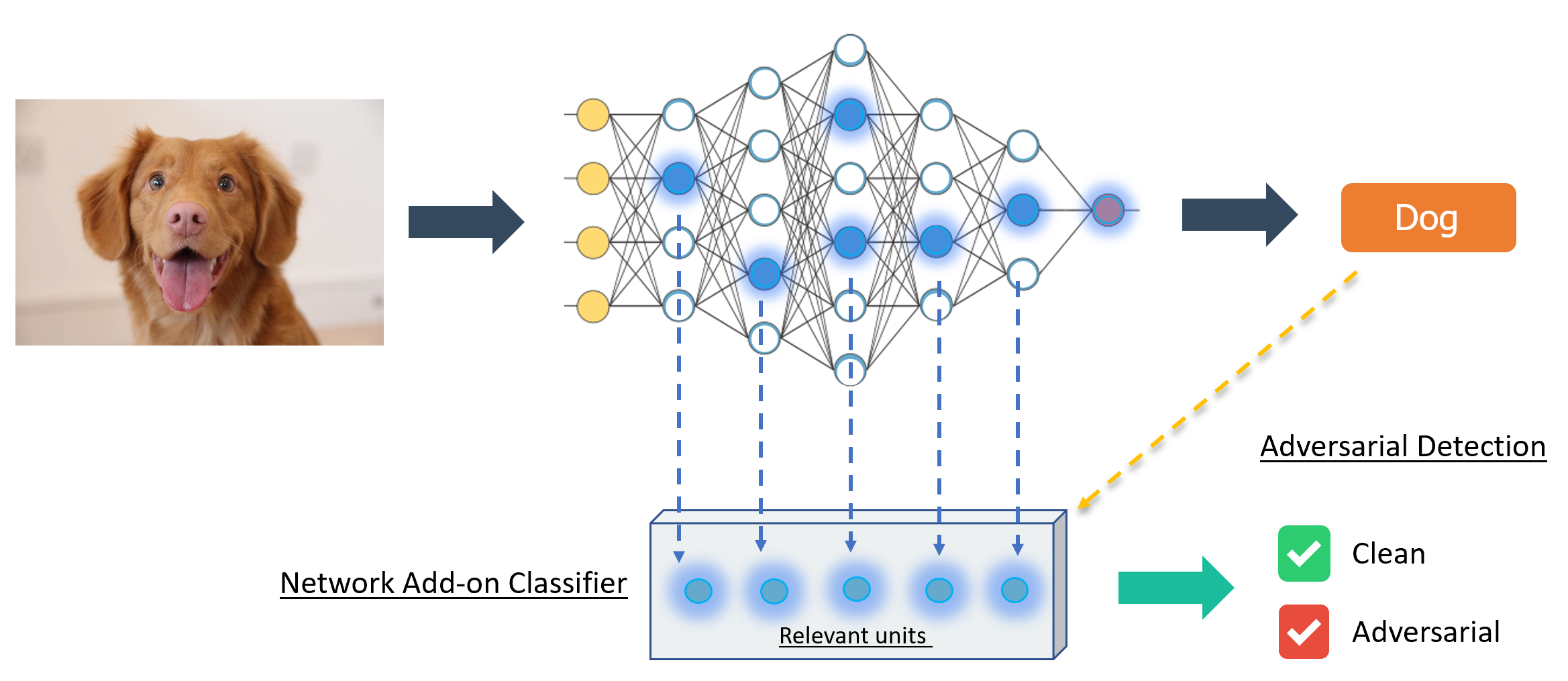}
  
  \caption{
  Proposed pipeline for adversarial attack detection.
  Detection is achieved by inspecting shared neurons that are relevant for the prediction of each class under clean and adversarial conditions.}
  \label{fig:schema}
\end{figure}

In order to verify this hypothesis, we propose an approach that assesses the relationship between the state of these sparse relevant neurons and the prediction given by the model. 
More specifically, given a pre-trained model, e.g. a convolutional neural network for image recognition, we identify internal neurons that are relevant for the classes predicted by the model~\cite{escorcia2015relationship,oramas2017visual}. Then, as depicted in Figure~\ref{fig:schema}, using the internal state of the identified neurons we train a classifier tasked with predicting whether the input provided to the model has been perturbed, i.e adversarially attacked. Following this idea, we perform a systematic analysis considering different factors, namely, the amount of selected neurons, the type of examples (original vs. perturbed) used when training the detectors, etc, that may have a significant role on the performance of the proposed method.

The contributions of this manuscript are two-fold:
i) A new perspective towards the detection of adversarial attacks driven by the inspection of relevant neurons in a deep neural network. ii) A light-weight method for the detection of adversarial attacks that has competitive performance w.r.t. similar state-of-the-art methods from the literature.

This paper is organized as follows: In Sec.~\ref{sec:relatedWork} we position our work w.r.t. existing methods in the literature.
Then, in Sec.~\ref{sec:observationMotivation}, we show preliminary tests further motivating the proposed method.
In Sec.~\ref{sec:proposedMethod} we describe the proposed method.
A validation, quantitative comparison w.r.t. existing method and a deeper inspection are conducted in Sec.~\ref{sec:evaluation}.
Finally, we draw conclusions in Sec.~\ref{sec:conclusion}.

\section{Related Work}
\label{sec:relatedWork}

The survey presented in \cite{akhtar2018threat} classifies defenses against Adversarial Attacks as: \textit{Complete Defense}, when there is a unique robust model that provides a correct prediction on any sample; \textit{Detection-Only Defense}, when an additional component verifies whether a sample is \textit{clean} or \textit{adversarial}; \textit{Modified Training/Input}, when defenses are based on modified datasets for training; \textit{Modified Network}, for strategies that rely on modified well-known network architectures; and \textit{Network Add-on}, for defense mechanisms outside the network. 
In order to test our hypothesis, we will leave the base network untouched. Hence, our study will be conducted following the \textit{Network Add-on} and \textit{Detection-only} approaches.  We propose to use a classifier to raise a flag for \textit{clean} or \textit{adversarial} samples.  Several authors have explored similar solutions  \cite{meng2017magnet,liang2017detecting,ma2019nic,xu2017feature}. Two closely related works will be described next.

Network Invariance Checking (NIC)~\cite{ma2019nic}, aims at identifying adversarial  samples by comparing which neurons are commonly activated when predicting \textit{clean} samples of a class. They refer to this concept as \textit{Provenance Invariant}, which is a similar situation to Flow Hijacking~\cite{abadi2009control} when an attack affects the  normal execution path of a computer program.  Additionally, in some cases Adversarial attacks provoke change on activation values in regularly activated neurons.  This is referred to as \textit{Value Invariant}.
The proposed detector is an ensemble add-on component of simpler models that collect activation values of subsequent layers inside a DNN. This detector searches for changes on \textit{Provenance} \&  \textit{Values} on neurons. Similar to ours, this is a \textit{Network add-on} defense based on neuron activation values. Different from NIC~\cite{ma2019nic} which uses activation values from \textit{all} layers; our method aims to spot specific neurons that are causing miss-classification of the DNN model. In this sense, our method reduced the complexity of detection.


The Feature Squeezing~\cite{xu2017feature} method for detecting attacks proposes the use of two additional models for classifying a feature-reduced version of the original input. Adversarial detection is achieved by analyzing the agreement/discrepancy between the prediction made by the original model and those made by the additional models.
Two Feature-Squeezing Methods were tested: \textit{Color-depth Reduction}  and   \textit{Spatial Smoothing}.
Following this idea, Feature-Squeezing reduces unnecessary feature space in input images, that can be used by adversarial attacks to craft samples \cite{xu2017feature}. This work is similar to ours in the sense that both are \textit{Detection-Only} \textit{Network Add-on} methods. 
Different from ours this method relies on manually deciding which feature compression technique works better for defending against a particular attack.

\section{Observations on Unit's Activation}
\label{sec:observationMotivation}
We refer in this work as unit/neuron to each filter/channel-wise responses of a convolutional layer.  As stated by \cite{rouhani2019safe,goodfellow2014explaining, tabacof2016exploring},  in a high n-dimensional feature space, features collected from neuron activations of adversarial samples reside in unstable and rarely explored regions near the boundaries. In this way, the vulnerability towards adversarial samples could be caused by a finite number of samples in a dataset or by classifiers that do not generalize well on boundaries \cite{rouhani2019safe}. 

Complementary to the above-mentioned, when visualizing the internal activations of clean and adversarial samples an intriguing situation occurs. We noticed that the activation vector of  adversarial samples, promoting the prediction of a given class, are pushed  towards the centroid of a cluster defined by clean samples of the corresponding class. 

\begin{figure}
  \centering
  \includegraphics[width=0.5\textwidth]{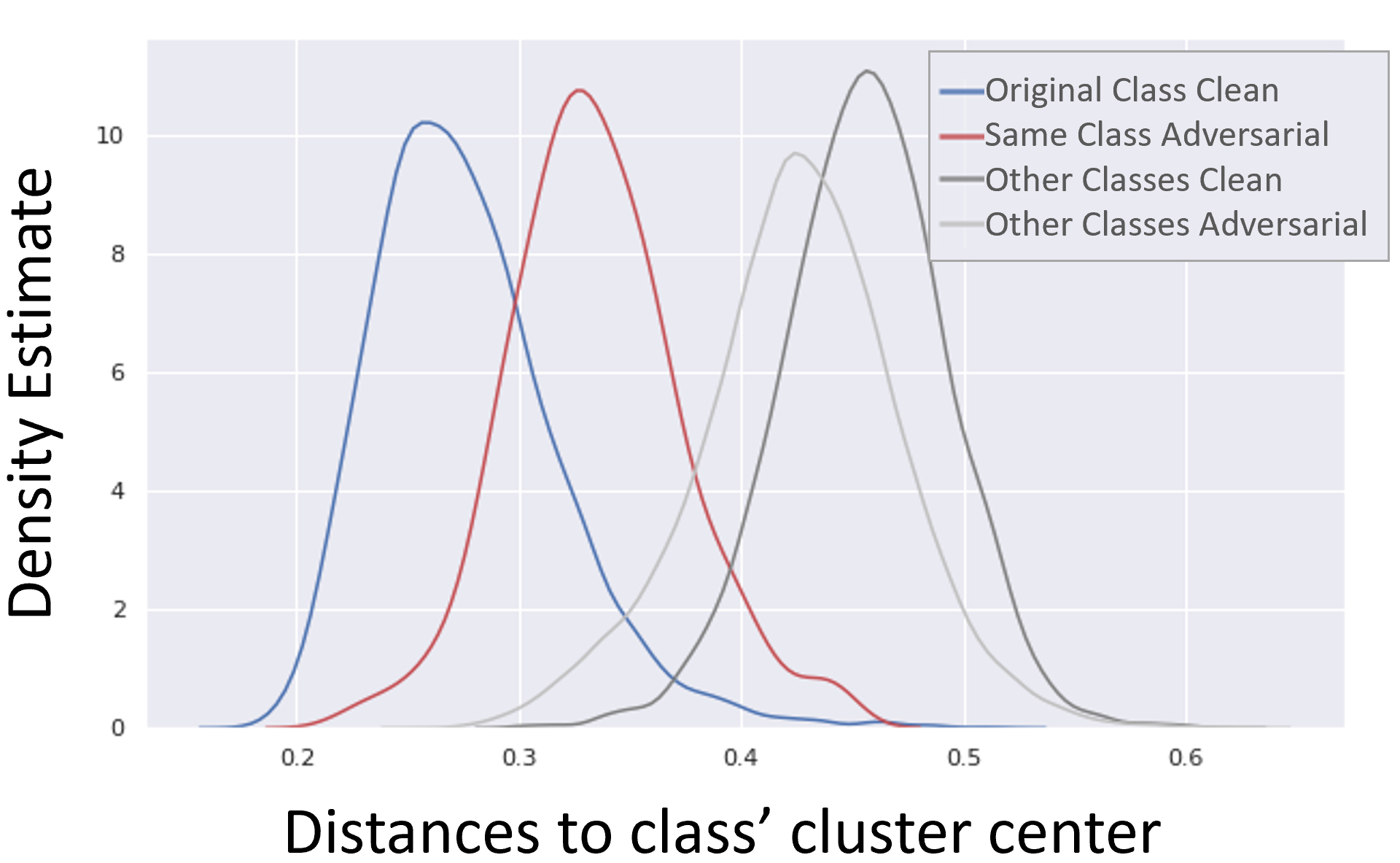}
  \caption{
  Probability density of distances, between activation vectors, to a given class cluster centroid. 
  These activations where collected from a DenseNet architecture trained to solve the classification problem from CIFAR10. Adversarial samples where collected by applying a JSMA attack. 
   We show all distances of each sample to their correspondent class's cluster centroid (\textit{Original class}), distances of the adversarial sample to the cluster centroid of the targeted class (\textit{Same class Adversarial}), and distances of all other samples of the dataset with a different class (\textit{Other Classes Clean} \& \textit{Other Classes Adversarial}). Results for all curves are aggregated for all classes.}
  \label{fig:distances1}
\end{figure}

This is observed when measuring the distances between the activation vectors from all the samples w.r.t. a cluster centroid defined by all the samples of a given class.
More specifically, we collect the activations from all the internal neurons and use them as feature vectors for clustering. Then, we measure the euclidean distance of each sample to the cluster centroid of a given class. 
In Figure \ref{fig:distances1} we can observe four curves depicting the probability density functions computed using the respective distances between clean and adversarial samples to the cluster centroid of a given class. The first curve represents the distance from each clean sample to its class cluster centroid. As expected examples belonging to a particular class will have the smallest distance of their corresponding centroid. The second curve represents distance of adversarial samples to the centroid of the targeted class. As can be seen, this curve in red has the closest distance to the attacked class compared with the remaining curves. The other two curves represent the same as above but in this case considering the rest of the classes. As observed the two curves depict a larger distance to the centroid. 

These plots were reconstructed using activation values of 1000 clean images and 1000 JSMA attack adversarial images of all classes from CIFAR10 training dataset. The neurons's activations were extracted from a Densenet model. This shows that adversarial samples are closer to samples of its target class than to those of other classes. In what follows we will aim at identifying the neurons that are responsible for the observed reduced distance.

\section{Proposed Method}
\label{sec:proposedMethod}

This section describes the proposed method for adversarial detection. The first part describes the sparse feature selection technique used as indicators for adversarial attack detection. Then, we describe a prototype system designed to evaluate our adversarial detection method.  


\subsection{Relevant unit selection}
\label{sec:relevantUnitDectection}
Deeper inspection of neurons/units has shown that there exists individual neurons that respond for a specific class or  subset of classes in a DNN \cite{zhou2018revisiting, escorcia2015relationship}. Thus,  we think it is important to conduct an analysis on the activation of neurons from clean and adversarial examples of each class. In order to detect which neurons are important for prediction on each of the classes of interest, we use the procedure for sparse unit selection described by \cite{oramas2017visual}. 

To execute this  procedure consider a network $f$ and $N$ training images. 
As a first step, we push every image through the network and for each image collect all the neuron activation values. For each layer, we compute the squared $L_2$ norm of each channel, thus producing a filter-level response. In this study we limit ourselves to only considering activations from convolutional and fully-connected layers. Then, we create a 1-dimensional descriptor by concatenating the filter-level responses from the different channels. Each resulting vector can be considered as a layer-specific descriptor. 
As a second step, each layer-specific descriptor from the previous step is divided by its $L_2$-norm. This helps  to compensate for the difference in length, i.e. number of channels, among different layers. 
As third step, we concatenate all the layer-specific descriptors to obtain a vector $x_i \in \mathbb{R}^{m}$. Repeat this procedure for all layers except for the last layer.  Here $m$ is defined by the number of all filter-wise response at all different layers. 
As fourth step, we construct a matrix $X \in \mathbb{R}^{m \times N}$ collecting all  $x_i \in \mathbb{R}^{m}$. In this case, the i-th image is represented by the column vector $x_i \in \mathbb{R}^{m}$, $N$ is the number of training images. 
Next, we construct a one-hot-encoded label matrix for all samples $L \in  \mathbb{R}^{C \times N}$. Where the i-th column vector $l \in \mathbb{R}^{C}$ of $L$ is a one-hot-encoded vector of the label of the i-th image. $C$ represents all possible classes on the selected dataset. 
Finally, the relevant neurons for each of the classes of interest are identified by solving the following equation:
\begin{equation}
    \label{eq:featureSelection}
    \begin{aligned}
    W^*=argmin_W \left\| X^TW - L^T \right\|^2_2 \qquad \\ \textrm{subject to:} \left\| w_j \right\|_1 \leq \mu, \ \forall_j=1..C 
    \end{aligned}
\end{equation}

The $\mu$ parameter controls the sparsity of the solution. The resulting optimization is formulated as a \textit{$\mu$-Lasso} problem which can be solved by efficiently using Spectral Gradient Projection \cite{mairal2014sparse,van2008probing}. According to \cite{oramas2017visual}, the \textit{$\mu$-Lasso} formulation is optimal for problems where exits a large number of activations compared to the number of available samples, which is the case of a Deep Neural Network. When the problem is solved the resulting matrix has  the form of $W^*=[w_1,w_2,..,w_c]$ where the i-th $w_i \in \mathbb{R}^{m} $ vector represents the relevance/importance of the unit/neuron/filter $1..m$ for the i-th class. The  $w_i \in \mathbb{R}^{m} $ is an sparse vector (most of values are zero); the sparsity is enforced by adding the constraint $\left\| w_j \right\|_1 \leq \mu$. 

The procedure for identification of relevant features is summarized in Algorithm~\ref{alg:sparseFeatureSelection}.

\begin{algorithm}
\caption{Sparse Relevant Unit Selection}
\label{alg:sparseFeatureSelection}
\begin{algorithmic}[1]

\STATE Compute all layer-specific descriptors in a Network $f$ on $N$ training images.
\STATE Divide each layer-specific descriptor by its $L_2$-norm.
\STATE Concatenate all the layer-specific descriptors to obtain a vector $x_i \in \mathbb{R}^{m}$. 
\STATE Construct a matrix $X \in \mathbb{R}^{m \times N}$ collecting all  $x_i \in \mathbb{R}^{m}$ from $N$ samples. 
\STATE Construct a one-hot-encoded label matrix for all samples $L \in  \mathbb{R}^{C \times N}$.
\STATE Solve $ W^*{=}argmin_W \left\| X^TW {-} L^T \right\|^2_2 , \left\| w_j \right\|_1 {\leq} \mu$
\end{algorithmic}
\end{algorithm} 

\begin{figure*}
\begin{center}
\includegraphics[width=1\textwidth]{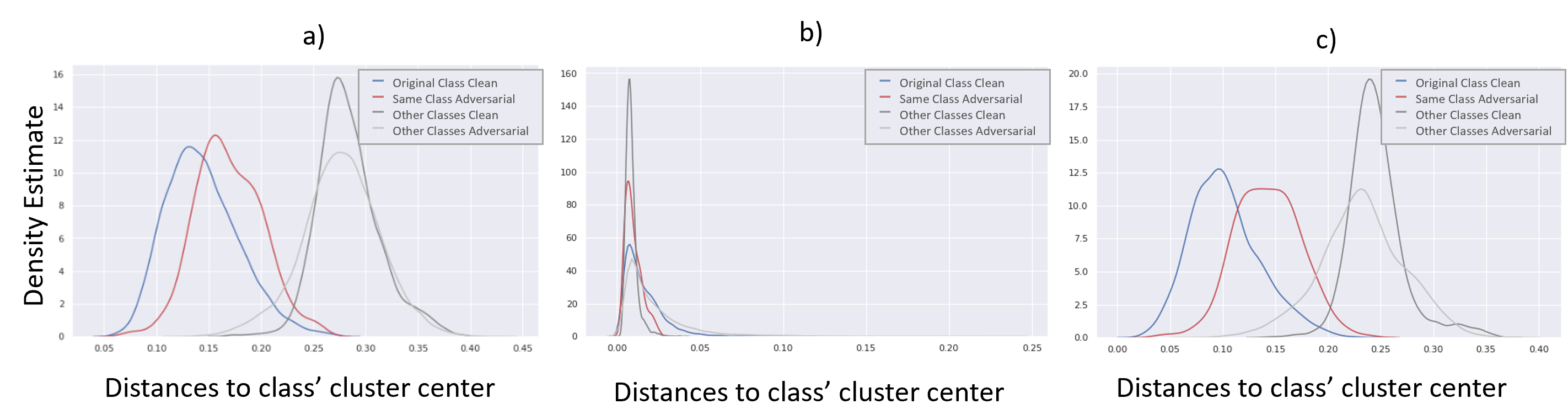}
\end{center}
   \caption{a) KDE plot of distances of relevant neuron's activation values to the corresponding cluster centroids of their classes. 
   b) KDE plot of distances of randomly selected activation values, this graphic was constructed for validation.
  c) KDE plot of distances of shared relevant neuron's activation values ($r_{shared-A}$) (between same class of adversarial and clean samples) to the corresponding cluster centroids of their clean classes.  All plots were generated using 1000 clean images of each class, and 1000 JSMA attack adversarial samples for each class of the CIFAR10 classes collected from DenseNet.   \textit{Original class} label represents all distances of each sample to their correspondent class's cluster centroid.  \textit{Same class Adversarial} represents distances of the adversarial sample to the cluster centroid of the targeted class. \textit{Other Classes Clean} and \textit{Other Classes Adversarial} represent distances of all other samples of the dataset with a different class. Results for all curves are aggregated for all classes.}
\label{fig:important}
\end{figure*}


\subsubsection{Sanity Check}
\label{sec:sanityCheck}
We perform the procedure above focusing on samples from one class at a time. Thus, as a result, for each class we obtain a set  \textit{$r_{clean-A}$ } of relevant neurons important for the prediction on clean samples from class $A$.
Similar to Section~\ref{sec:observationMotivation}, we explored what happens when measuring distances of samples to their cluster centroids. This time however by only taking activation values from relevant neurons for the class. 
The idea is to perform a sanity check to verify to what extent, the observations made in Section~\ref{sec:observationMotivation} still hold.
Figure~\ref{fig:important}.a depicts a density functions of the distances in analogous conditions as in Figure~\ref{fig:distances1} for relevant neurons. As can be seen, similar to previous observations, the curve corresponding to adversarial samples is pushed towards the distribution of clean samples of the class they target. 
 One can think that activation values from relevant neurons contain valuable information that can be useful for adversarial detection.  
 In order to further verify this idea, in Figure~\ref{fig:important}.b we plot a similar curve, but this time by considering randomly selected neurons for each class. As can be noticed, randomly selected neuron activations do not provide sufficient discriminative information to help distinguish between clear/attacked examples or between examples of different classes.


\subsubsection{Sharing Relevant Neurons between Clean and Adversarial Samples}
\label{sec:sharingUnits}

Additionally, if we perform relevant neuron selection procedure for adversarial images targeting a particular class A, we obtain a set $r_{adv-A}$ of relevant neurons related to class A. 
Then, if we intercept both sets of relevant neurons for clean and adversarial samples, we obtain a new set $r_{shared-A}= r_{clean-A} \cap r_{adv-A}$. This new set  $r_{shared-A}$ contains neurons that are relevant for class $A$ in both conditions (\textit{clean} and \textit{adversarial}). This set contains an small number of neurons (20-40 in the case of CIFAR10 and DeseNet). 
Constructing the probabilistic density function from the euclidean distances to the class's cluster centroids, we obtain the Figure~\ref{fig:important}.c. As can be seen, we reach to a similar observation as before. The adversarial features distances are closer to the cluster centroid of the clean class they target. One can think that this small set still contains valuable information that can be used for the detection of adversarial examples. 

The above-mentioned observations indicate that it could be useful to explore the detection of adversarial examples at a class level. Hence, our updated hypothesis is: \textit{Given a set $r_{clean-A}$ of relevant neurons important for the prediction on clean samples of class $A$, an adversarial attack adds small perturbation on samples from other classes, thus this perturbation changes the activation patterns of subset of neurons $r_{clean-A}$ in order to obtain a miss-classification as class $A$.}


\subsection{Adversarial Attack Detection}

In order to test the detection of adversarial attacks based on the state of the relevant neurons, we follow the pipeline presented in Figure \ref{fig:schema}. The proposed detection system is characterized as an \textit{Network Add-on} and \textit{Detection only} solution. Under this schema we propose the use an add-on classifier to work along neural network models. 

The proposed classifier is aimed to classify each image sample as \textit{clear} or \textit{adversarial}. It uses activation values of relevant neurons from set $r_{shared-A}$ described in the last section; paired with the prediction given by the network. 
This classifier should use the network prediction to decide which activation values are discriminative enough to assist the detection. 
Those activation values must come from relevant neurons for the predicted class $A$. To feed the classifier we define a feature vector which considers the relevant neurons from all the classes of interest. When constructing the feature vector, one can set to zero (0) part of the vector that does not correspond relevant neurons of the predicted class. By considering all the relevant neurons from all the classes we are capable of defining fixed-length feature vectors which is desirable when working with standard classifiers.

\vspace{4mm}
\noindent{\textbf{Training:}}

The training process of the system is summarized in Algorithm \ref{alg:training}.
In practice the training process of the proposed method is composed by the following steps: 
As first step, we generate adversarial samples of any attack technique of interest for all samples of class $A$. Class $A$ denotes any of the classes belonging to the dataset. As second step, we perform selection of the relevant neurons of class $A$ on clean samples (Sec.~\ref{sec:relevantUnitDectection})
Likewise, as third step, we perform selection of relevant neurons on adversarial (attacked) samples that provoked a prediction of class $A$. As fourth step, we identify shared relevant neurons $r_{shared-A}$ for clean class $A$ and adversarial class $A$. As final step, we train an \textit{adversarial classifier} using clean and adversarial images using the activation values $r_{shared-A}$ from the previous step. 

\begin{algorithm}
\caption{Training procedure for the proposed method} Perform selection of relevant unit  of class $A$ on adversarial samples.

\label{alg:training}
\begin{algorithmic}[1]
\STATE Generate adversarial samples for all samples of class $A$.
\STATE Perform selection of relevant unit of class $A$ on clean samples.
\STATE Perform selection of relevant unit  of class  $A$ on adversarial samples.
\STATE Identify Shared Relevant Neurons $r_{shared-A}$ for clean class $A$ and adversarial class $A$.
\STATE Train a simple model (\textit{adversarial classifier}) using clean and adversarial images using activation values from  units on $r_{shared-A}$. 
\end{algorithmic}
\end{algorithm} 

We use as \textit{adversarial classifier} a simple random forest with 1000 trees  without hyper-parameter tuning.

\vspace{4mm}
\noindent{\textbf{Testing:}}
The testing procedure for adversarial detection is rather simple, and it does not add considerable overhead in computation time. As a first step, for each sample passed through the network we collect activation values from the $r_{shared-A}$ set of relevant neurons for the predicted class $A$. Then, we construct the feature vector with those activation values,and feed it to the classifier to obtain a  \textit{clean} / \textit{adversarial} prediction.

\section{Evaluation}
\label{sec:evaluation}

We conduct three experiments to assess the strengths and weaknesses of the proposed method. The first experiment (Sec.~\ref{sec:validationExp}) aims to validate our detection method against three adversarial attack methods. 
This experiment is used to answer our research question. 
The second experiment (Sec.~\ref{sec:comparisonExp}) quantitatively positions our method w.r.t. related state-of-the-art detectors. 
The third experiment (Sec.~\ref{sec:gradcam}) is intended to qualitatively asses our method. 

\textbf{Datasets}
Following the standard protocol~\cite{akhtar2018threat} for adversarial attack detection, we run experiments in both the MNIST~\cite{lecun1998gradient} and CIFAR10~\cite{krizhevsky2009learning} datasets. MNIST contains 28x28 gray-scale images depict digit classes. It contains 60K images for training and 10K for testing. CIFAR10 contains 32x32 color images depicting 10 object classes. It contains 50K images for training and 10k for validation.

\begin{figure}
  \centering
  \includegraphics[width=0.5\textwidth]{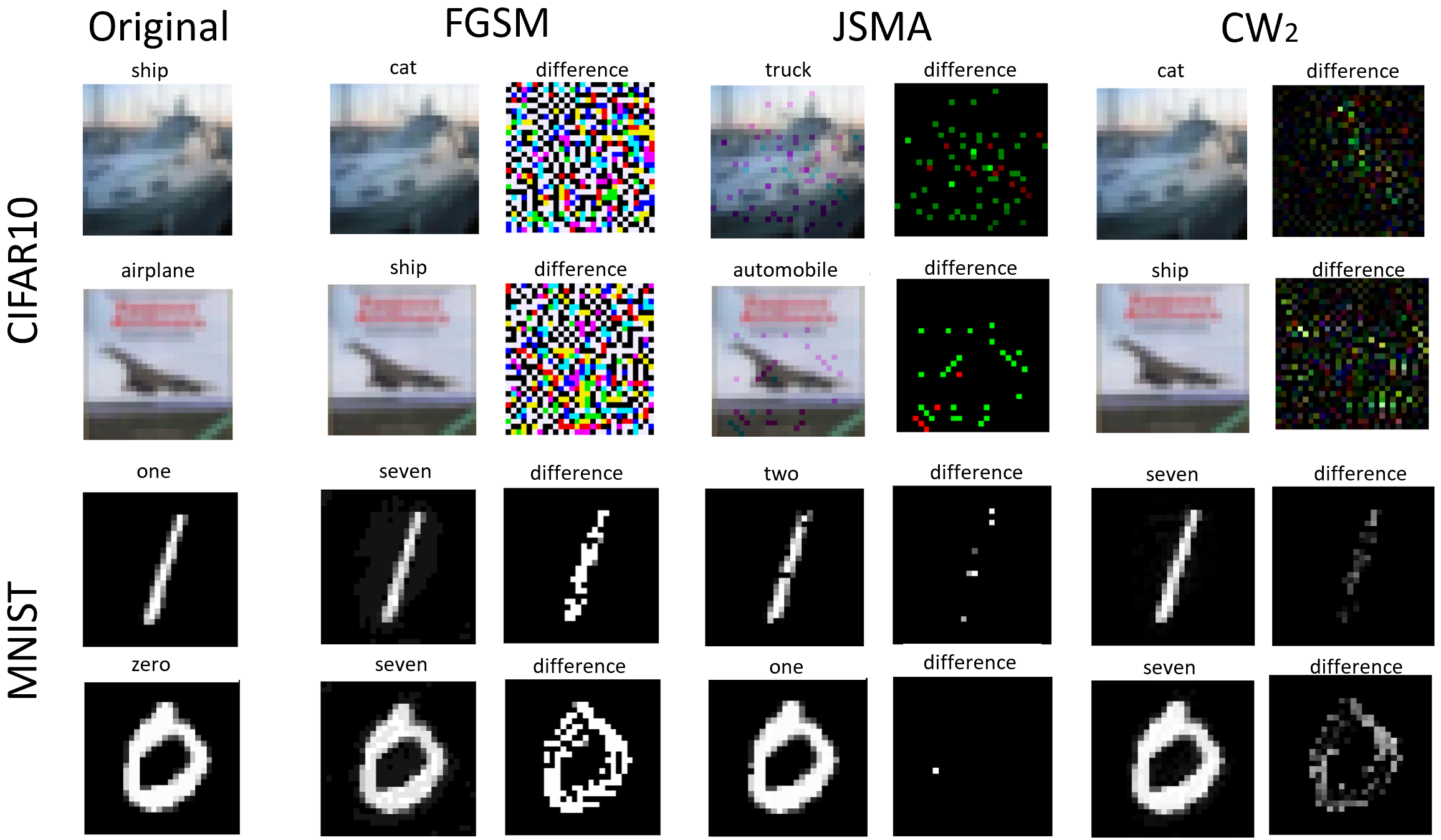}
  \caption{ Adversarial samples generated for FGSM, JSMA and CW2 attacks
on the CIFAR10 and MNIST datasets using the LeNet and DenseNet architectures. 
Each column contains generated adversarial images and a difference images showing how the  pixels are altered the in original image in order to
achieve miss-classification. 
Predicted labels are shown at the top
of each sample.
}
  \label{fig:samples}
\end{figure}

\textbf{Adversarial Attacks:}
Throughout our evaluation, we measure the effectivity of the proposed method at detecting modified examples from three different adversarial attack methods, namely FGSM \cite{goodfellow2014explaining}, JSMA \cite{papernot2016limitations} and $CW_2$ \cite{carlini2017towards}. 
Adversarial samples were generated using Foolbox Library~\cite{rauber2017foolbox} using parameters described in each section.
Fig.~\ref{fig:samples} depicts samples of adversarial generated adversarial images for CIFAR10 and MNIST datasets.

\begin{table*}[]
\centering
\begin{tabular}{l|l|ccc|ccc|}
\hline
\multirow{2}{*}{\textbf{Selected Neurons}} & \multirow{2}{*}{ \textbf{Attack}} &
\multicolumn{3}{c|}{\textbf{With Predicted Label}} & \multicolumn{3}{c|}{ \textbf{W/O Predicted Label}} \\ \cline{3-8} 
 & & \textbf{$\mu{=}100$} & \textbf{$\mu{=}20$} & \textbf{$\mu{=}10$} & 
  \textbf{$\mu{=}100$} & \textbf{$\mu{=}20$} & \textbf{$\mu{=}10$} \\ \hline
 \multirow{3}{*}{\begin{tabular}[c]{@{}l@{}}Shared Relevant \\ Neurons \\ ($ r_{clean-A} \cap r_{adv-A}$)\end{tabular}}
 
 & FGSM & 0,969 & 0,982 & 0,983 & 0,977 & 0,966 & 0,912 \\ 
 & JSMA & 0,984 & 0,974 & 0,966 & 0,953 & 0,912 & 0,832 \\ 
 & C\&W2 & 0,980 & 0,979 & 0,974 & 0,969 & 0,963 & 0,964 \\ \hline
 
 \multirow{3}{*}{\begin{tabular}[c]{@{}l@{}}Only Relevant from \\ Clean Class\\ ($r_{clean-A}$)\end{tabular}}
 & FGSM & 0,982 & 0,982 & 0,985 & 0,977 & 0,976 & 0,963 \\
 & JSMA & 0,981 & 0,977 & 0,972 & 0,953 & 0,942 & 0,841 \\ 
 & C\&W2 & 0,978 & 0,979 & 0,978 & 0,966 & 0,969 & 0,957 \\ \hline

\multirow{3}{*}{\begin{tabular}[c]{@{}l@{}}Relevant non-shared\\ from both classes \\ ($r_{clean-A} \triangle r_{adv-A}'$)\end{tabular}}
& FGSM & 0,981 & 0,980 & 0,981 & 0,977 & 0,977 & 0,971 \\
& JSMA & 0,973 & 0,978 & 0,974 & 0,954 & 0,945 & 0,916 \\ 
& C\&W2 & 0,974 & 0,977 & 0,978 & 0,965 & 0,969 & 0,958 \\ \hline

\end{tabular}
\caption{Adversarial detection accuracy of the proposed method 
on a LeNet architecture trained on the MNIST dataset.}
\label{tab:exp1_MNIST}
\end{table*}


\begin{table*}[]
\centering
\begin{tabular}{l |l |c c c | c c c | }
\hline

\multirow{2}{*}{\textbf{Selected Neurons}} & \multirow{2}{*}{ \textbf{Attack}} &
\multicolumn{3}{c|}{\textbf{With Predicted Label}} & \multicolumn{3}{c|}{ \textbf{W/O Predicted Label}} \\ \cline{3-8} 
 & & \textbf{$\mu{=}100$} & \textbf{$\mu{=}20$} & \textbf{$\mu{=}10$} & 
  \textbf{$\mu{=}100$} & \textbf{$\mu{=}20$} & \textbf{$\mu{=}10$} \\ \hline

  \multirow{3}{*}{ \begin{tabular}[c]{@{}l@{}}Shared Relevant \\ Neurons \\ ($ r_{clean-A} \cap r_{adv-A}$)\end{tabular}}
  & FGSM & 0,944 & 0,956 & 0,954 & 0,844 & 0,864 & 0,864 \\ \cline{2-8} 
  & JSMA & 0,946 & 0,932 & 0,926 & 0,929 & 0,871 & 0,864 \\ \cline{2-8} 
  & C\&W2 & 0,892 & 0,881 & 0,879 & 0,862 & 0,706 & 0,721 \\ \hline

  \multirow{3}{*}{ \begin{tabular}[c]{@{}l@{}}Only Relevant \\ from Clean Class\\ ($r_{clean-A}$)\end{tabular}} 
  & FGSM & 0,937 & 0,952 & 0,953 & 0,810 & 0,854 & 0,818 \\ \cline{2-8} 
  & JSMA & 0,945 & 0,939 & 0,934 & 0,934 & 0,879 & 0,875 \\ \cline{2-8} 
  & C\&W2 & 0,893 & 0,875 & 0,878 & 0,802 & 0,703 & 0,698 \\ \hline  

  \multirow{3}{*}{ \begin{tabular}[c]{@{}l@{}}Relevant non-shared\\ from both classes \\ ($r_{clean-A} \triangle r_{adv-A}'$)\end{tabular}}
  & FGSM & 0,879 & 0,900 & 0,915 & 0,788 & 0,849 & 0,860 \\ \cline{2-8} 
  & JSMA & 0,924 & 0,931 & 0,926 & 0,931 & 0,928 & 0,932 \\ \cline{2-8} 
   & C\&W2 & 0,848 & 0,882 & 0,873 & 0,823 & 0,871 & 0,880 \\ \hline
\end{tabular}
\caption{Adversarial detection accuracy of the proposed method 
on a DenseNet architecture trained on the CIFAR10 dataset.}
\label{tab:exp1_CIFAR10}
\end{table*}

\subsection{Exp-1: Validation}
\label{sec:validationExp}

This experiment aims at validating the capability of our method for the detection adversarial attacks
In this experiment we consider a LeNet architecture  \cite{lecun1998gradient} using ReLu Layers; trained in MNIST  \cite{lecun1998gradient}, and a DenseNet architecture \cite{huang2017densely}; trained in CIFAR10  \cite{krizhevsky2009learning}, as the models to be attacked. Adversarial attack parameters were the default parameters encountered in Foolbox~\cite{rauber2017foolbox}. These parameters restrict perturbation on crafted adversarial samples.

We consider here three variants of the proposed method which consider different relevant neurons. The first variant considers only the shared neurons ($r_{shared}$), which are relevant neurons that are commonly excited when predicting samples in \textit{clean} and \textit{adversarial} condition at the same class.
The second variant considers only the set of neurons ($r_{clean-A}$) that are relevant for a particular class in clean condition only. 
The third variant considers relevant neurons that are not shared between clean and adversarial examples. It can be though as symmetric difference of sets of relevant neurons in clean and adversarial condition $r_{clean-A} \triangle r_{adv-A}$. 
In addition, as part of this experiment we assess the effect of the sparsity parameter $\mu$ (Sec.~\ref{sec:proposedMethod}) for the selection of relevant neurons. Towards this goal, we conduct experiments with multiple $\mu$ values $(\mu{=}10, 20, 100)$.

The ultimate task of our detector is to predict whether a sample is \textit{clean} or \textit{adversarial} by looking at a vector that collects activation of relevant neurons of all classes.
We gauge our detector in two different conditions: \textit{With-Label} and \textit{Without-Label}. In the \textit{Without-Label} condition, the classifier receives information of all relevant neurons from all classes in the dataset without discrimination. This means that the classifier has to learn to discriminate between the classes of interest, initially addressed by the model, and clean/adversarial classes that characterize the examples. 
In the \textit{With-Label} condition, we utilize the predicted label to pass just the information from only the neurons that belong to the predicted class. This is achieved by zero-ing all activation values that belong to relevant neurons of other classes. 

For this experiment we select 1000 samples of each class on training dataset which are used to generate adversarial samples for its corresponding dataset/model pair. These samples are then used to train the Random-Forest classifier used for detection. Performance on attack detection is done from unseen images using all test-set from the datasets: 10K original/clean images and corresponding 10K adversarial samples.

\begin{figure}
 
  \includegraphics[width=0.5\textwidth]{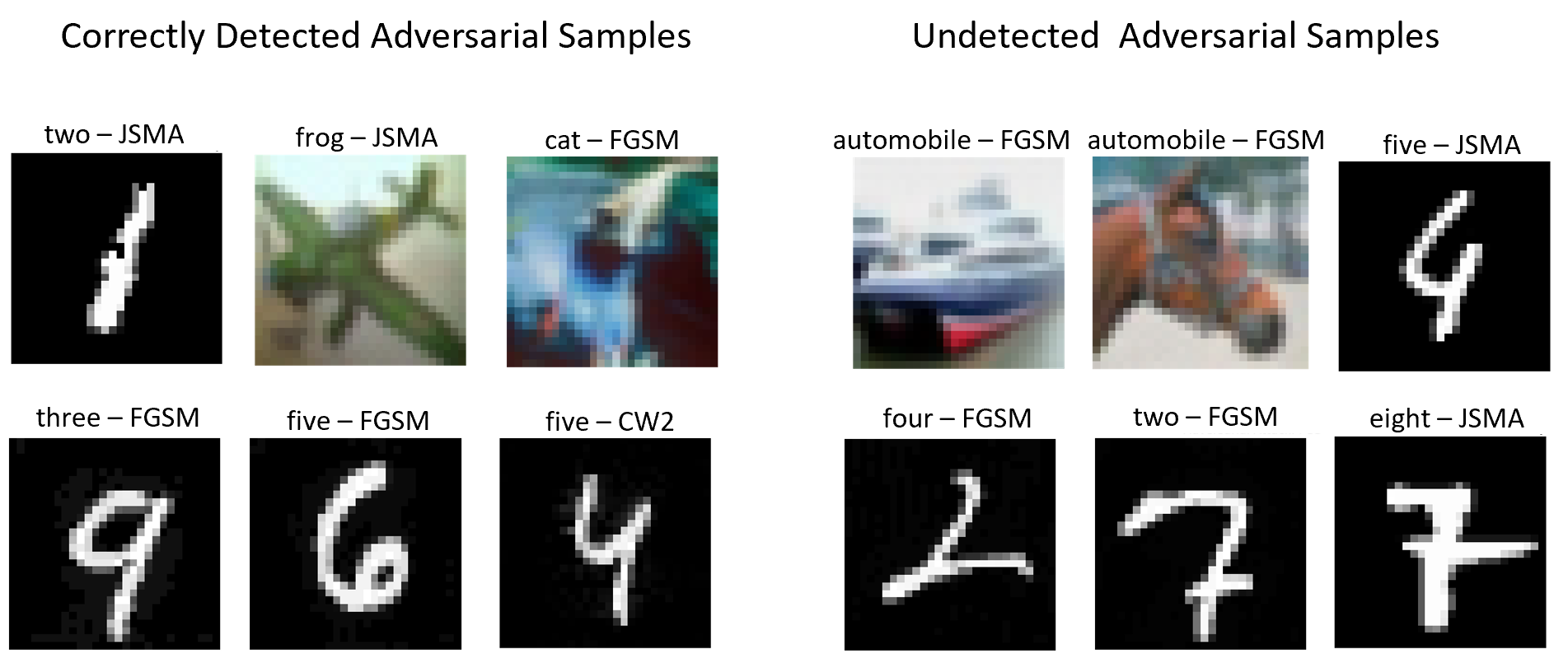}
  \caption{ Examples of correctly detected adversarial samples  vs undetected adversarial samples by the proposed method. On the top of each image we provide the predicted class and the attack method used to craft the adversarial sample}
  \label{fig:detected}
\end{figure}

\textbf{Results:}
Results from this experiment are reported in a Table~\ref{tab:exp1_MNIST} and Table~\ref{tab:exp1_CIFAR10} for the MNIST and CIFAR10 datasets, respectively.

The total number of neurons from the convolutional layers belonging to LeNet model were 244; while for the DenseNet model there were 3272 units. The $\mu$ value controls the sparsity of the solution found when identifying the relevant neurons. The higher the value of $\mu$, the more relevant neurons are selected for each class.

As can be noticed, the configuration/variant used to define the vector of relevant neurons that is fed to the classifier has relatively low effect in the detection accuracy.
This is important since, despite using less information in the $r_{shared}$ variant, our method was still relatively accurate at detection. 
In addition, for the case of the $r_{clean}$ variant, there is no need to perform identification of relevant neurons from adversarial samples, which leads to reduced computation costs.
In the MNIST dataset our proposed method is able to detect any adversarial attack with accuracy higher than 0.96 on all \textit{With-Label} conditions and most of \textit{Without-Label} condition. 
For the CIFAR10 dataset, in the $r_{shared}$ condition results are different. For FGSM and JSMA attack methods, the detection  accuracy is higher than 0.926; whereas for the CW$L_2$ attack method the reported accuracy is slightly lower (0.892 at best). 

Regarding the effect of the sparsity parameter $\mu$, it can be noticed that even with the lowest $\mu{=}10$ value, which implies a very small number of neurons selected for each class (i.e. 5.3\% to 18.3\% of total neurons from LeNet, trained in MNIST; 0.3\% to 0.9\% from DenseNet, trained in CIFAR10), our method achieves detection with decent accuracy on both datasets.  We also show that $\mu$ can be considered as a tuning parameter that may lead to better detection. For the settings considered in our experiments the highest difference in accuracy introduced by a change of this parameter was around 3\%.

Figure~\ref{fig:detected} depicts samples of correctly detected adversarial samples and adversarial samples that bypassed our method. It can be noticed that the undetected samples from the MNIST dataset share considerable similarities with the predicted adversarial class (digit '2' compared to predicted digit '4', or digit '7' compared to predicted digit '2' ) and are even challenging for humans. However, from the undetected samples from the CIFAR10 dataset it is not trivial to draw a conclusion on potential sources of error.

\textbf{Discussion:}
Our research question was: \textit{Is it possible to detect unseen adversarial samples of a given class by inspecting neurons that are usually over-excited by an adversary sample?}. Experiment 1 shows that it is possible to detect adversarial attacks with decent accuracy of any class by inspecting a small set of shared neurons $r_{shared}$. These nuerons are the ones relevant for the prediction provided by the network in clean and adversarial images (less than 46\% of total neurons in a LeNet model trained on MNIST; less than 2\% in a DenseNet model trained for all the classes from CIFAR10) using $\mu{=}10$.

 Detection of adversarial attacks is performed with high accuracy for weak attacks, while for strong attacks accuracy is still decent. Our detector also performs very well reporting a low number of false positives. We believe that the $CW_2$  detection accuracy was slightly lower on CIFAR10 due to the fact that this attack is one of the strongest from the literature. Future research should explore this limitation with other datasets and attack parameters and probably considering more relevant neurons. In comparison with NIC~\cite{ma2019nic}, we achieve slightly less, but still comparable, detection accuracy. It is worth reiterating that this accuracy is achieved by inspecting up to 50\% of neurons in LeNet and less than 2\% on DenseNet, while NIC uses all neurons and even multiple detectors to achieve the reported accuracy. This means that our method has the potential to have better scalability for more complex models and larger datasets. 

In this way we show that our approach reduces the search space for detection of adversarial attacks inside a DNN. 


\begin{table}[]
\centering
\resizebox{1\columnwidth}{!}{
\begin{tabular}{lcccc}
\hline

\textbf{Attack} & \textbf{Ours$(\mu{=}10)$} & \textbf{Ours$(\mu{=}20)$} & \textbf{NIC~\cite{ma2019nic} } & \textbf{Fea. Squeez.~\cite{xu2017feature}} \\ 
\midrule
FGSM & \textbf{0,985} & 0,980 & 1,000 & 1,000 \\ 
JSMA & 0,975 & \textbf{0,990} & 1,000 & 1,000 \\ 
CWL2 & 0,986 & \textbf{0,993} & 1,000 & 1,000 \\ \hline

\end{tabular}
}
\caption{Comparison with state-of-the-art detectors on adversarial detection on the MNIST dataset. Best results in bold.}
\label{tab:exp2_MNIST}
\end{table}

\begin{table}[]
\centering
\resizebox{1\columnwidth}{!}{
\begin{tabular}{lcccc}
\hline

\textbf{Attack} & \textbf{Ours$(\mu{=}10)$} & \textbf{Ours$(\mu{=}20)$} & \textbf{NIC~\cite{ma2019nic} } & \textbf{Fea. Squeez.~\cite{xu2017feature}} \\ 
\midrule
FGSM & \textbf{0,940} & 0,940 & 1,000 & 0,210 \\ 
JSMA & 0,941 & \textbf{0,951} & 0,960 & 0,840 \\ 
CWL2 & 0,805 & \textbf{0,820} & 0,940 & 1,000 \\ \hline

\end{tabular}
}
\caption{Comparison with state-of-the-art detectors on adversarial detection on the CIFAR10 dataset. Best results in bold.}
\label{tab:exp2_CIFAR10}
\end{table}


\begin{figure*}[ht!]
\centering
  \includegraphics[width=0.99\textwidth]{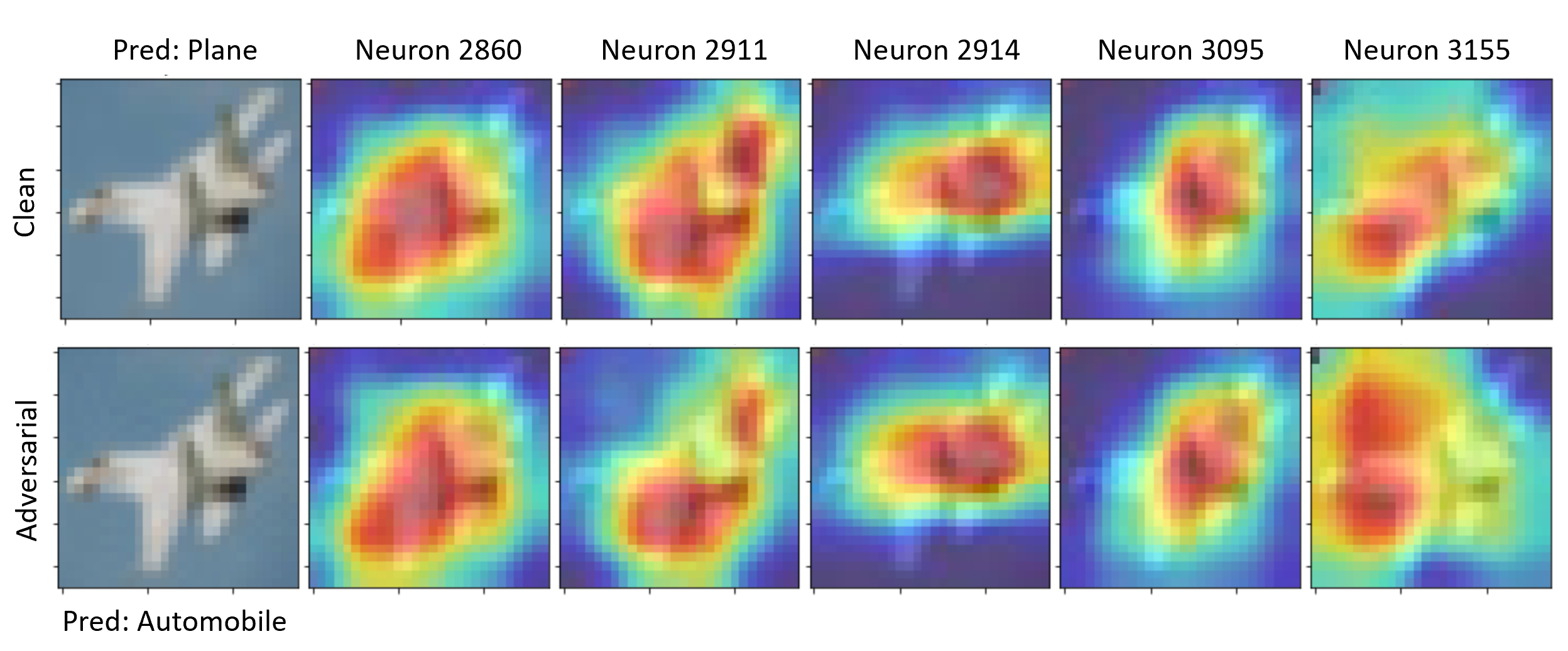} \includegraphics[width=0.99\textwidth]{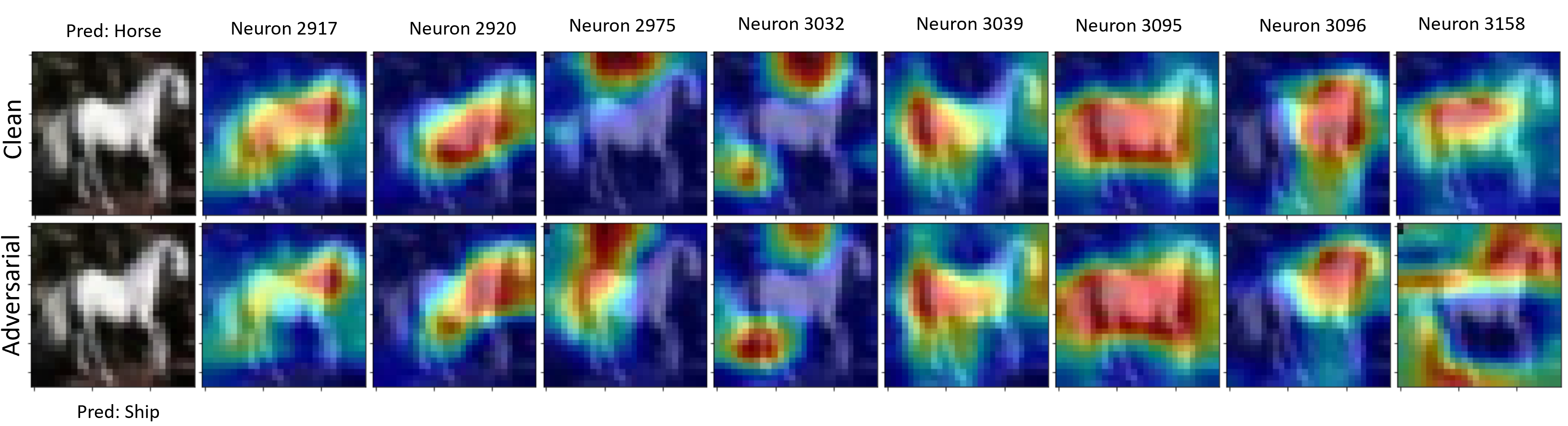}
    \caption{GradCAM visualization of each shared-relevant neurons from classes in CIFAR10. For each example we show the visualization of it correctly predicted class plane/horse (top) and its adversarially predicted class ship/automobile (bottom).}
  \label{fig:shared1}
\end{figure*}


\subsection{Exp-2: Comparison w.r.t. existing methods}
\label{sec:comparisonExp}
This experiment was carried out in order to assess the performance of the proposed method with respect to other state-of-the-art methods for adversarial attack detection. Here we consider NIC~\cite{ma2019nic} and Feature Squeezing~\cite{xu2017feature}. Similar to the last experiment, adversarial attacks tested in this comparison are: FGSM \cite{goodfellow2014explaining}, JSMA \cite{papernot2016limitations} and $CW_2$ \cite{carlini2017towards}. We adopted the adversarial sample generation strategy, parameters, and sample size found in their corresponding studies in order to make the results comparable. For FGSM we utilized \textit{untargeted attack} and parameter $max\_epsilon{=}0.3$. For JSMA and $CW_2$ we used the \textit{targeted-attack} or \textit{white-box attack} strategy: The targeted class selection was decided using the formula $T{=} L{+}1\  mod /C $; where $L$ is the original class of the sample image and \textit{C} is 10 for MNIST and CIFAR10.  Parameters for JSMA were  $theta{=}1.0,max_iter{=}51$, whereas for $CW_2$ were $binary\_search\_steps {=} 9, max\_iterations{=} 1000,$ $learning\_rate {=} 1e-2, confidence {=} 0, initial\_const {=}1e-3$. All parameters are the same/equivalent to those found in the Cleverhans Library~\cite{papernot2018cleverhans} as considered in NIC and Feature-Squeezing studies. For this evaluation we selected the first 100 samples that were correctly predicted and found correctly generated adversarial samples from the test-set for the three attacks, as performed in both studies. In this experiment we maintain the use of the CIFAR10 and MNIST datasets. For CIFAR10 we use the same DenseNet architecture from the previous experiment. Whereas on the MNIST dataset we used the "MNIST-model" from \cite{carlini_nn} described in the experimental setup. We tested two variants of our method with sparsity parameter $\mu{=}10 $and $\mu{=}20$. Training our adversarial classifier was performed using adversarial samples generated on the previous experiment. In comparison, adversarial attack samples for the evaluation in Experiment 2 are stronger attacks than those applied on samples of Experiment 1. 

\textbf{Results:}
We report adversarial detection results from this experiment in Table~\ref{tab:exp2_MNIST} and \ref{tab:exp2_CIFAR10}. 
Detection accuracy is near perfect for the MNIST-model for all attacks with low false-positive rate (3\% for JSMA attack). NIC and Feature Squeezing achieve perfect accuracy. However these methods are significantly more complex than ours. Moreover, those models were trained  on adversarial samples with stronger attack parameters. On the CIFAR10 dataset, our results are still positive, for FSGM and JSMA we achieve high detection accuracy. Similarly, NIC achieves high accuracy for FGSM while Feature Squeezing reported only 0.20. The same case is found for JSMA, our method reported 0.951 as best accuracy while Feature Squeezing just achieves 0.84. For $CW_2$ our best accuracy is 0.82 while the other works obtain higher accuracy than ours. For JSMA and FGSM our method is again depicting low false-positive rate. For $CW_2$ false positive-rate is higher (around 14\%) than on other attacks.

\textbf{Discussion:}
Based on the observations made from Table~\ref{tab:exp2_MNIST} and \ref{tab:exp2_CIFAR10} we can state that our method achieves comparable results with state-of-the-art detectors~\cite{ma2019nic, xu2017feature}. Moreover, the implementation of our technique is simpler than the analyzed detectors; since it focuses on a lower number of neurons inside a DNN in order to recognize adversarial samples. 
More specifically 60 neurons in total for all classes in CIFAR10/DenseNet for sparsity parameter $\mu{=}10$ (see Table \ref{tab:exp1_CIFAR10}). These means that we use less than 2\% of the total number of neurons. In contrast, NIC~\cite{ma2019nic} uses all 3172 neurons for CIFAR10 and a large quantity of models in parallel to achieve comparable performance on adversarial attack detection.   
Furthermore, it is worth to mention that training of our adversarial classifier was performed with weaker adversarial samples. This indicates that our method is able to recognize stronger adversarial inputs without being trained for them. Carlini-Wagner $L_2$ ($CW_2$) attack in CIFAR10 was the most difficult attack for our detector. Our method only achieves around 82\% of detection on this attack. It is important to explore solutions for this limitation in future work. A possible solution could be to train on $CW_2$ with different perturbations. Despite of that this is a positive result considering that it only took an small set of neurons (between 20 to 46) to recognize adversarial samples. We also show that it is possible to tune the accuracy of detection by changing the sparsity of relevant unit selection.

\subsection{Exp-3: Inspecting Relevant Neurons}
\label{sec:gradcam}

\begin{figure}
  \includegraphics[width=0.5\textwidth]{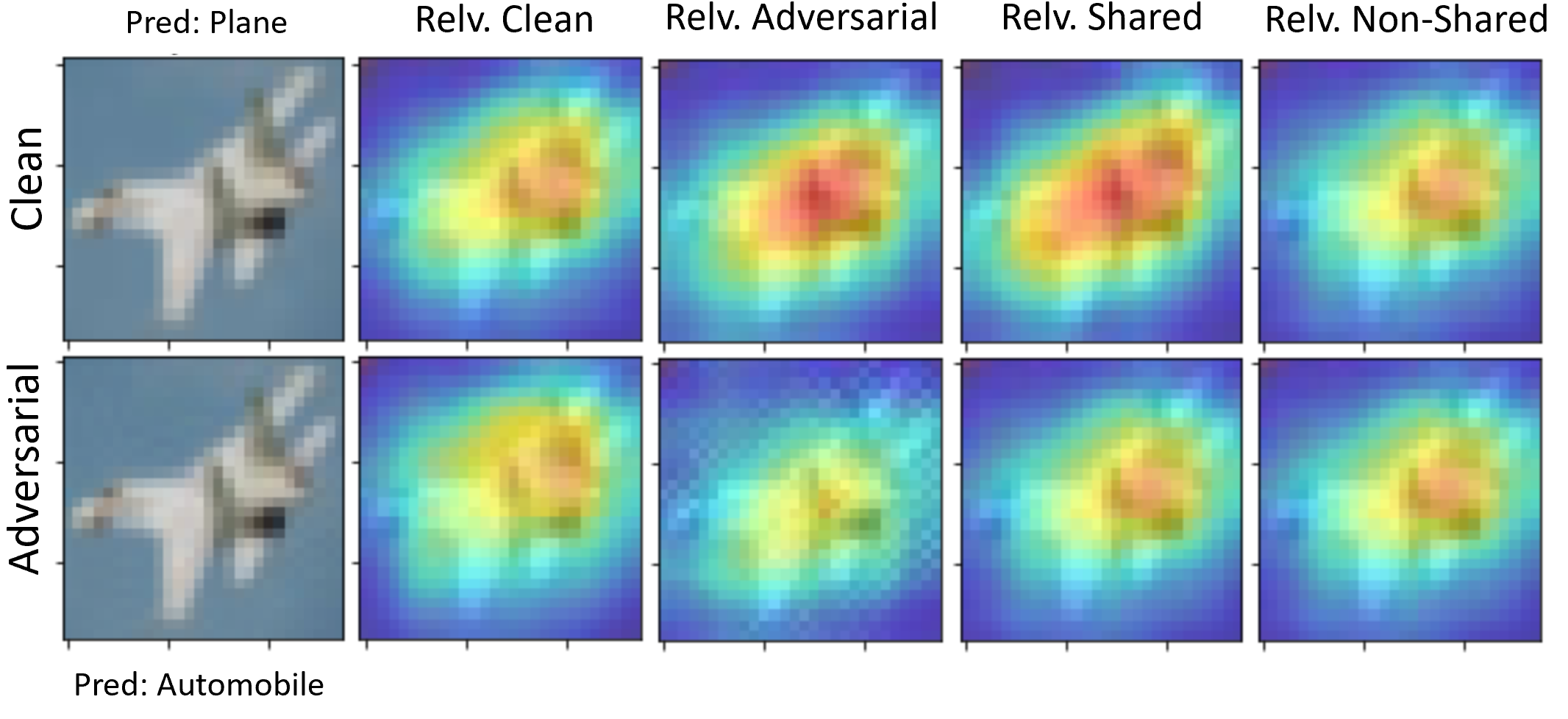}
  \includegraphics[width=0.5\textwidth]{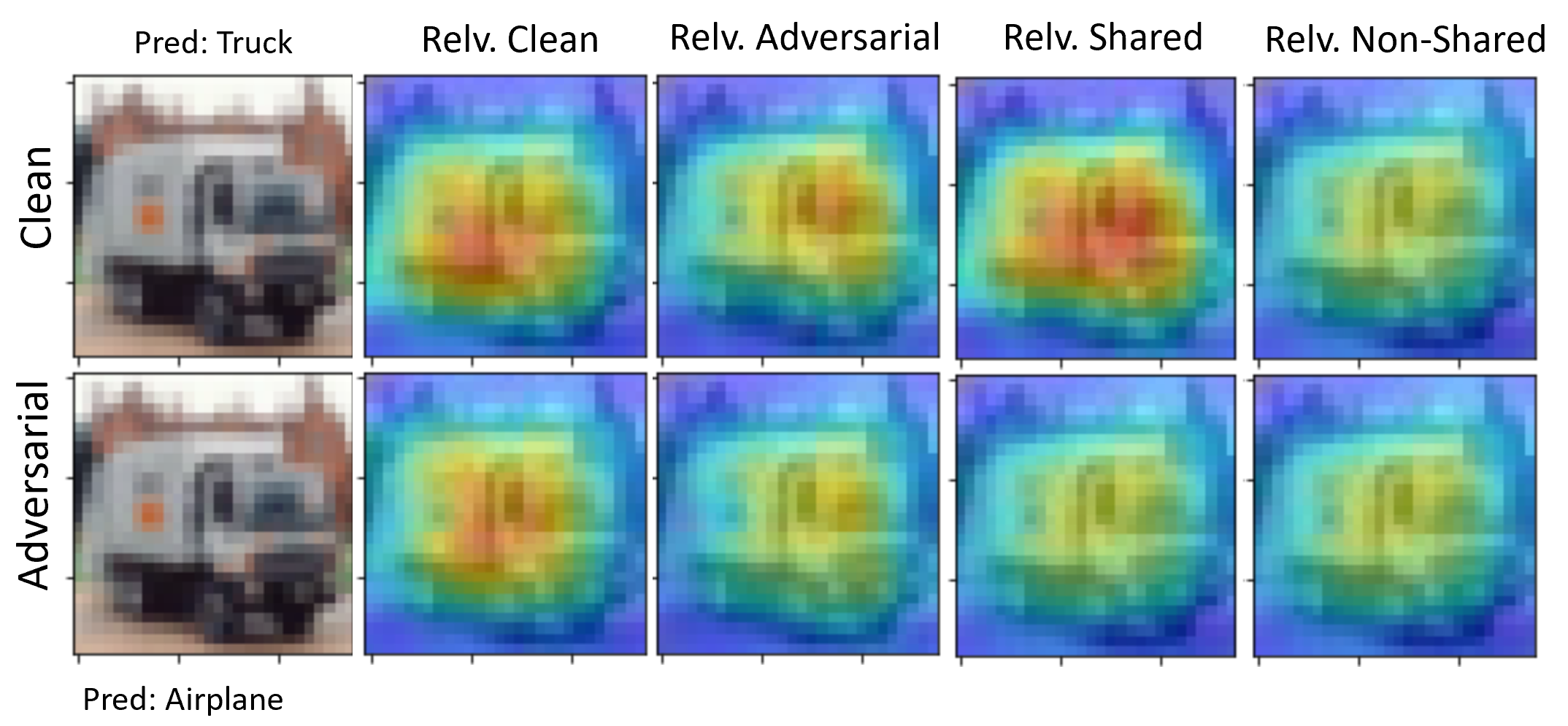}
  
  \caption{ Combined GradCAM visualization of different sets of relevant neurons in examples from CIFAR10
  under different grouping settings.
  For each example we show the visualization in clean (top) and under adversarial attack conditions (bottom). 
  }
  \label{fig:merged1}
\end{figure}

Here we aim at getting an insight on how the subtle perturbations 
present in the adversarial samples change the behaviour of the 
identified relevant neurons.
Towards this goal we generate heatmap visualizations 
using GradCam \cite{selvaraju2017grad}.  
for each of the relevant neurons and combinations of them.
Visualizations of isolated neurons is achieved by locating the layer/filter where the relevant neurons are located and setting the weight of other filters in that layer to zero within the GradCAM formulation. This effectively isolates the effect of the neuron to be inspected. 
We compare these visualization when observing the original and its corresponding attacked sample in Fig.~\ref{fig:shared1}.
We also combined the heatmaps from the isolated relevant neurons in order to visualize what different sets of neurons are sensing. Fig.~\ref{fig:merged1} 
shows combined visualizations for:  1) relevant Neurons identified for the predicted class in clean conditions, 2) relevant Neurons the same class in adversarial conditions, 3) shared-relevant neurons for the same class in adversarial and clean conditions. 4) Neurons that were relevant for a class in one condition (clean or adversarial) but not in both.


\textbf{Discussion:}
As can be seen in Fig.~\ref{fig:shared1},  the visualizations differ when comparing clean and adversarial samples. In some cases, e.g. Neuron-2914 (top) and Neuron-3039 (bottom), differences are rather subtle. 
In other cases, e.g. Neuron-3155 (top) or in Neuron-3158 (bottom), there is a clear difference on what the neurons are sensing despite the similarity of their inputs. 
In the combined visualization (Fig.~\ref{fig:merged1}), shared-relevant neurons show greater differences than other sets of neurons.  In general, we can observe that shared relevant neurons exhibit greater changes when comparing a given image in clean and adversarial conditions.
This shows how the subtle changes introduced by the adversarial attack 
make the model focus its attention in different regions of the input image.


\section{Limitations and Future Work}
The evaluation of our method has been focused on an stand-alone attacker, i.e. without knowledge of the detector. Future efforts should explore adaptive attacker settings, since it has been demonstrated that attacker can develop countermeasures for defenses~\cite{akhtar2018threat,carlini2017magnet}. Randomization techniques should also be explored to develop robust defense in this method as other detectors proposed \cite{xu2017feature,su2019one}.
Finally, experiments on more complex models and datasets should be conducted since our method seems to be suitable to work on models with large number of neurons.

\section{Conclusion}
\label{sec:conclusion}

We presented a method to achieve adversarial attack detection at class level by analyzing the relationship between activation of relevant neurons and the predictions made by the model. 
Our experiments show that an adversarial attack usually change the state of a small set of neurons that are relevant to the prediction of a targeted class.
Despite its relative simplicity, our results suggests that our method is capable of detecting adversarial attacks with comparable performance to the state-of-the-art detectors. Moreover, this is achieved by only considering a relatively small set of neurons from the model being protected.

\textbf{Acknowledgements:} This work was partially supported by the UAntwerp BOF DOCPRO4-NZ project (ID 41612) "Multimodal Relational Interpretation for Deep Models", and the FWO Fundamental Project (G0A4720N) "Design and Interpret: A New Framework for Explainable Artificial Intelligence".

{\small
\bibliographystyle{ieee_fullname}
\bibliography{egbib}

\begin{thebibliography}{10}\itemsep=-1pt

\bibitem{abadi2009control}
Mart{\'\i}n Abadi, Mihai Budiu, {\'U}lfar Erlingsson, and Jay Ligatti.
\newblock Control-flow integrity principles, implementations, and applications.
\newblock {\em ACM Transactions on Information and System Security (TISSEC)},
  13(1):4, 2009.

\bibitem{akhtar2018threat}
Naveed Akhtar and Ajmal Mian.
\newblock Threat of adversarial attacks on deep learning in computer vision: A
  survey.
\newblock {\em IEEE Access}, 6:14410--14430, 2018.

\bibitem{carion2020endtoend}
Nicolas Carion, Francisco Massa, Gabriel Synnaeve, Nicolas Usunier, Alexander
  Kirillov, and Sergey Zagoruyko.
\newblock End-to-end object detection with transformers, 2020.

\bibitem{carlini_nn}
Nicholar Carlini.
\newblock Robust evasion attacks against neural network to find adversarial
  examples.
\newblock \url{https://github.com/carlini/nn_robust_attacks}, 2016.

\bibitem{carlini2017magnet}
Nicholas Carlini and David Wagner.
\newblock Magnet and" efficient defenses against adversarial attacks" are not
  robust to adversarial examples.
\newblock {\em arXiv preprint arXiv:1711.08478}, 2017.

\bibitem{carlini2017towards}
Nicholas Carlini and David Wagner.
\newblock Towards evaluating the robustness of neural networks.
\newblock In {\em 2017 IEEE Symposium on Security and Privacy (SP)}, pages
  39--57. IEEE, 2017.

\bibitem{escorcia2015relationship}
Victor Escorcia, Juan Carlos~Niebles, and Bernard Ghanem.
\newblock On the relationship between visual attributes and convolutional
  networks.
\newblock In {\em Proceedings of the IEEE Conference on Computer Vision and
  Pattern Recognition}, pages 1256--1264, 2015.

\bibitem{frankleLotteryTicketICLR19}
Jonathan Frankle and Michael Carbin.
\newblock The lottery ticket hypothesis: Training pruned neural networks.
\newblock In {\em ICLR}, 2019.

\bibitem{goodfellow2014explaining}
Ian~J Goodfellow, Jonathon Shlens, and Christian Szegedy.
\newblock Explaining and harnessing adversarial examples.
\newblock {\em arXiv preprint arXiv:1412.6572}, 2014.

\bibitem{deepCompressionICLR16}
Song Han, Huizi Mao, and William~J. Dally.
\newblock Deep compression: Compressing deep neural network with pruning,
  trained quantization and huffman coding.
\newblock In Yoshua Bengio and Yann LeCun, editors, {\em ICLR}, 2016.

\bibitem{HeResNetCVPR15}
Kaiming He, Xiangyu Zhang, Shaoqing Ren, and Jian Sun.
\newblock In {\em Deep Residual Learning for Image Recognition}, 2016.

\bibitem{hintonDistillingNIPS15}
Geoffrey Hinton, Oriol Vinyals, and Jeff Dean.
\newblock Distilling the knowledge in a neural network.
\newblock In {\em NeurIPS Workshops}, 2014.

\bibitem{huang2017densely}
Gao Huang, Zhuang Liu, Laurens Van Der~Maaten, and Kilian~Q Weinberger.
\newblock Densely connected convolutional networks.
\newblock In {\em Proceedings of the IEEE conference on computer vision and
  pattern recognition}, pages 4700--4708, 2017.

\bibitem{krizhevsky2009learning}
Alex Krizhevsky, Geoffrey Hinton, et~al.
\newblock Learning multiple layers of features from tiny images.
\newblock Technical report, Citeseer, 2009.

\bibitem{lecun1998gradient}
Yann LeCun, L{\'e}on Bottou, Yoshua Bengio, Patrick Haffner, et~al.
\newblock Gradient-based learning applied to document recognition.
\newblock {\em Proceedings of the IEEE}, 86(11):2278--2324, 1998.

\bibitem{lewisBARTACL20}
Mike Lewis, Yinhan Liu, Naman Goyal, Marjan Ghazvininejad, Abdelrahman Mohamed,
  Omer Levy, Veselin Stoyanov, and Luke Zettlemoyer.
\newblock {BART}: Denoising sequence-to-sequence pre-training for natural
  language generation, translation, and comprehension.
\newblock 2020.

\bibitem{liang2017detecting}
Bin Liang, Hongcheng Li, Miaoqiang Su, Xirong Li, Wenchang Shi, and Xiaofeng
  Wang.
\newblock Detecting adversarial image examples in deep networks with adaptive
  noise reduction.
\newblock {\em arXiv preprint arXiv:1705.08378}, 2017.

\bibitem{ma2019nic}
Shiqing Ma, Yingqi Liu, Guanhong Tao, Wen-Chuan Lee, and Xiangyu Zhang.
\newblock Nic: Detecting adversarial samples with neural network invariant
  checking.
\newblock In {\em NDSS}, 2019.

\bibitem{mairal2014sparse}
Julien Mairal, Francis Bach, Jean Ponce, et~al.
\newblock Sparse modeling for image and vision processing.
\newblock {\em Foundations and Trends{\textregistered} in Computer Graphics and
  Vision}, 8(2-3):85--283, 2014.

\bibitem{meng2017magnet}
Dongyu Meng and Hao Chen.
\newblock Magnet: a two-pronged defense against adversarial examples.
\newblock In {\em Proceedings of the 2017 ACM SIGSAC Conference on Computer and
  Communications Security}, pages 135--147. ACM, 2017.

\bibitem{moosavi2016deepfool}
Seyed-Mohsen Moosavi-Dezfooli, Alhussein Fawzi, and Pascal Frossard.
\newblock Deepfool: a simple and accurate method to fool deep neural networks.
\newblock In {\em Proceedings of the IEEE conference on computer vision and
  pattern recognition}, pages 2574--2582, 2016.

\bibitem{DNNSpeechRecognition19}
A.~B. {Nassif}, I. {Shahin}, I. {Attili}, M. {Azzeh}, and K. {Shaalan}.
\newblock Speech recognition using deep neural networks: A systematic review.
\newblock {\em IEEE Access}, 7:19143--19165, 2019.

\bibitem{oramas2017visual}
Jose Oramas, Kaili Wang, and Tinne Tuytelaars.
\newblock Visual explanation by interpretation: Improving visual feedback
  capabilities of deep neural networks, 2019.

\bibitem{papernot2018cleverhans}
Nicolas Papernot, Fartash Faghri, Nicholas Carlini, Ian Goodfellow, Reuben
  Feinman, Alexey Kurakin, Cihang Xie, Yash Sharma, Tom Brown, Aurko Roy,
  Alexander Matyasko, Vahid Behzadan, Karen Hambardzumyan, Zhishuai Zhang,
  Yi-Lin Juang, Zhi Li, Ryan Sheatsley, Abhibhav Garg, Jonathan Uesato, Willi
  Gierke, Yinpeng Dong, David Berthelot, Paul Hendricks, Jonas Rauber, and
  Rujun Long.
\newblock Technical report on the cleverhans v2.1.0 adversarial examples
  library.
\newblock {\em arXiv preprint arXiv:1610.00768}, 2018.

\bibitem{papernot2016limitations}
Nicolas Papernot, Patrick McDaniel, Somesh Jha, Matt Fredrikson, Z~Berkay
  Celik, and Ananthram Swami.
\newblock The limitations of deep learning in adversarial settings.
\newblock In {\em 2016 IEEE European Symposium on Security and Privacy
  (EuroS\&P)}, pages 372--387. IEEE, 2016.

\bibitem{rauber2017foolbox}
Jonas Rauber, Wieland Brendel, and Matthias Bethge.
\newblock Foolbox: A python toolbox to benchmark the robustness of machine
  learning models.
\newblock {\em arXiv preprint arXiv:1707.04131}, 2017.

\bibitem{rouhani2019safe}
BD Rouhani, M Samragh, T Javidi, and F Koushanfar.
\newblock Safe machine learning and defeating adversarial attacks. ieee secur
  priv. 2019; 17 (2): 31-28, 2019.

\bibitem{selvaraju2017grad}
Ramprasaath~R Selvaraju, Michael Cogswell, Abhishek Das, Ramakrishna Vedantam,
  Devi Parikh, and Dhruv Batra.
\newblock Grad-cam: Visual explanations from deep networks via gradient-based
  localization.
\newblock In {\em Proceedings of the IEEE international conference on computer
  vision}, pages 618--626, 2017.

\bibitem{su2019one}
Jiawei Su, Danilo~Vasconcellos Vargas, and Kouichi Sakurai.
\newblock One pixel attack for fooling deep neural networks.
\newblock {\em IEEE Transactions on Evolutionary Computation}, 2019.

\bibitem{szegedy2013intriguing}
Christian Szegedy, Wojciech Zaremba, Ilya Sutskever, Joan Bruna, Dumitru Erhan,
  Ian Goodfellow, and Rob Fergus.
\newblock Intriguing properties of neural networks.
\newblock {\em arXiv preprint arXiv:1312.6199}, 2013.

\bibitem{tabacof2016exploring}
Pedro Tabacof and Eduardo Valle.
\newblock Exploring the space of adversarial images.
\newblock In {\em 2016 International Joint Conference on Neural Networks
  (IJCNN)}, pages 426--433. IEEE, 2016.

\bibitem{van2008probing}
Ewout Van Den~Berg and Michael~P Friedlander.
\newblock Probing the pareto frontier for basis pursuit solutions.
\newblock {\em SIAM Journal on Scientific Computing}, 31(2):890--912, 2008.

\bibitem{xu2017feature}
Weilin Xu, David Evans, and Yanjun Qi.
\newblock Feature squeezing: Detecting adversarial examples in deep neural
  networks.
\newblock {\em arXiv preprint arXiv:1704.01155}, 2017.

\bibitem{zhou2018revisiting}
Bolei Zhou, Yiyou Sun, David Bau, and Antonio Torralba.
\newblock Revisiting the importance of individual units in cnns via ablation.
\newblock {\em arXiv preprint arXiv:1806.02891}, 2018.

\end{thebibliography}
}

\end{document}